\documentclass[journal]{IEEEtran}
\usepackage[colorlinks,urlcolor=blue,linkcolor=blue,citecolor=blue]{hyperref}
\usepackage{amsmath,amssymb,amsfonts}
\usepackage{algorithmic}
\usepackage{graphicx}
\usepackage{textcomp}
\usepackage{color,array}
\usepackage{multirow}
\usepackage{multicol}
\usepackage{graphicx}
\usepackage{caption}
\usepackage{footnote}
\usepackage{float}
\usepackage{tabularx}
\usepackage{booktabs}

\usepackage[caption=false]{subfig}
\usepackage{array}
\newcolumntype{x}[1]{>{\centering\arraybackslash}m{#1}}
\usepackage{scalerel}
\usepackage{tikz}

\usepackage{lipsum}

\usepackage{graphicx}
\usepackage{multirow}
\usepackage{subfig}
\usepackage{makecell}
\usepackage[utf8]{inputenc}

\usepackage{xcolor}

\definecolor{RB1}{HTML}{55ff00}
\definecolor{RB2}{HTML}{ff0000}
\definecolor{RB3}{HTML}{000cff}
\definecolor{RB4}{HTML}{ffff00}
\definecolor{RB5}{HTML}{147d7d}
\definecolor{RB6}{HTML}{aaaaff}
\definecolor{RB7}{HTML}{aa557f}
\definecolor{RB8}{HTML}{55aaff}
\definecolor{RB9}{HTML}{55ffff}
\definecolor{RB10}{HTML}{ef07ff}
\definecolor{LB1+2}{HTML}{ff8c39}
\definecolor{LB3}{HTML}{550000}
\definecolor{LB4}{HTML}{155b64}
\definecolor{LB5}{HTML}{ff5500}
\definecolor{LB6}{HTML}{aab600}
\definecolor{LB7+8}{HTML}{5e6284}
\definecolor{LB9}{HTML}{93d4ff}
\definecolor{LB10}{HTML}{da42ff}

\begin{document}

\title{Structure and position aware graph neural network for airway labeling}

\author{Weiyi Xie, \and
        Colin Jacobs, \and
        Jean-Paul~Charbonnier, \and
        Bram van Ginneken
\thanks{This work was supported by the Dutch Lung Foundation under the project 5.1.17.171. We acknowledge the COPDGene  Study (ancillary study  ANC-337) for providing the data used. COPDGene is funded by Award Number U01 HL089897 and Award Number U01 HL089856 from the National Heart, Lung, and Blood Institute. The content is solely the responsibility of the authors. It does not necessarily represent the official views of the National Heart, Lung, and Blood Institute of the National Institutes of  Health. (Corresponding author: Weiyi Xie. e-mail: Weiyi.Xie@radboudumc.nl).}%
\thanks{Weiyi Xie and Colin Jacobs and Bram van Ginneken are with the Diagnostic Image Analysis Group, Department of Medical Imaging, Radboudumc, 6525 GA Nijmegen, The Netherlands (weiyi.xie@radboudumc.nl; colin.jacobs@radboudumc.nl; bram.vanginneken@radboudumc.nl).}
\thanks{Jean-Paul Charbonnier is with Thirona, 6525 EC Nijmegen, The Netherlands (jeanpaulcharbonnier@thirona.eu).}} 

\markboth{}
{W. Xie \MakeLowercase{\textit{et al.}}: Structure and position aware graph neural network for airway labeling}

\maketitle
\begin{abstract}
We present a novel graph-based approach for labeling the anatomical branches of a given airway tree segmentation. The proposed method formulates airway labeling as a branch classification problem in the airway tree graph, where branch features are extracted using convolutional neural networks (CNN) and enriched using graph neural networks. Our graph neural network is structure-aware by having each node aggregate information from its local neighbors and position-aware by encoding node positions in the graph.  

We evaluated the proposed method on 220 airway trees from subjects with various severity stages of Chronic Obstructive Pulmonary Disease (COPD). The results demonstrate that our approach is computationally efficient and significantly improves branch classification performance than the baseline method. The overall average accuracy of our method reaches 91.18\% for labeling all 18 segmental airway branches, compared to 83.83\% obtained by the standard CNN method. We published our source code at https://github.com/DIAGNijmegen/spgnn. The proposed algorithm is also publicly available at https://grand-challenge.org/algorithms/airway-anatomical-labeling/.
\end{abstract}
\begin{IEEEkeywords}
Airway Labeling \and Graph Neural Networks \and Convolutional Neural Networks \and Chronic Obstructive Pulmonary Disease.
\end{IEEEkeywords}

\section{Introduction}
\IEEEPARstart{C}{hronic} obstructive pulmonary disease (COPD) is one of the most prevalent lung diseases and a leading cause of chronic morbidity and mortality worldwide \cite{adeloye2015global}. Airway narrowing and remodeling are typical COPD characteristics. Therefore, assessing airway remodeling is essential for evaluating disease severity and progression. CT imaging is an excellent tool for in vivo quantitative airway analysis. Such an analysis can be performed efficiently when an automatic airway segmentation is available and is often applied regionally for specific anatomical branches. An automated airway labeling can expedite such a process. Airway labeling is also useful to plan bronchoscopic interventions. We therefore aimed to develop an automatic algorithm for anatomical airway labeling given a pre-extracted airway tree. 

Airway labeling was traditionally performed by matching an unlabeled tree to a pre-labeled tree \cite{kitaoka2002automated,tschirren2005matching}, where correspondences between the two trees were defined using visual and topological features extracted on branches or branch points. Most methods exploited prior knowledge about the airway anatomy. For example, Tschirren et al.~\cite{tschirren2005matching} searched for an optimal match between two trees as a maximum clique on association graphs based on segment lengths, spatial orientations, and angle differences between branch segments to measure associations between branch points. Other methods based on matching trees involved geodesic distances in a tree space \cite{feragen2011,feragen2012,feragen2014}. Two trees are adjacent in tree space with similar visual features and common topology. Beyond matching-based approaches, supervised machine learning methods proposed to model a probability distribution of branch labels given their features on training data. By assuming features of different anatomical branches are independently Gaussian distributed, van Ginneken et al.~\cite{van2008robust} labeled airways by learning Gaussian distributions using branch features such as orientation, average radius, and the angle relative to the parent. The authors also exploited a graph-based topological assumption that a branch's labeling decision is conditioned on its parent's predicted label. Other supervised machine learning methods for airway labeling include a hidden Markov tree model \cite{ross2014airway}, branch classifiers \cite{mori2009automated} based on AdaBoost, and KNN based appearance models \cite{lo2011bottom}.\par
Airway labeling is challenging because airway tree topology varies substantially across subjects. Moreover, segmented trees may have missing or spurious branches due to imaging noise, and their structures may also be affected by pathological changes. Due to these challenges, traditional airway labeling methods may not generalize well on unseen trees if they rely heavily on hand-craft features or specific rules derived from airway anatomy. As data-driven approaches, convolution neural networks (CNNs) and graph neural networks (GNNs) offer to extract powerful visual features and encode airway anatomy holistically. A popular regime in this topic is to combine CNNs with the power of GNNs, where CNNs extract visual features for object representation, and GNNs optimize representations by learning structural relations between objects according to their connections in the graph. A method called deep vessel segmentation \cite{shin2019deep} incorporated a GNN into a CNN architecture to jointly exploit both visual appearances and local structures for segmenting vessels in retinal images and coronary artery X-ray angiography. For airway segmentation in CT images, Juarez et al.~\cite{juarez2019joint} trained a GNN to capture airway connectivity using dense features from a CNN, and both networks were trained end-to-end, but with a memory spending limit. A very recent work \cite{tan2021sgnet} proposed to segment and label airways simultaneously by jointly training CNNs and GNNs. In this method, dense features from a CNN network were used to extract features for branch points, and these features were augmented in a GNN by considering the information within 2-hop neighbors for each branch point. \par
The existing methods that combine CNNs with GNNs have three limitations. First, most approaches adopted standard GNNs, i.e., graph convolution networks \cite{kipf2016semi} in \cite{juarez2019joint,tan2021sgnet} and graph attention networks \cite{velivckovic2017graph} in \cite{shin2019deep}. Standard GNNs are only structure-aware \cite{you2019position}. Their expressive power is limited by the 1-Weisfeiler-Lehman (WL) \cite{li2020distance} graph isomorphism test, meaning that two nodes with similar features and neighborhoods are hardly distinguishable. This structure-aware property may not be sufficient for node classification problems in graphs exhibiting symmetric structures, i.e., two nodes are located in different parts of a graph but with the same neighboring structure. Intuitively, adding positional information may be beneficial for identifying nodes with similar neighbors but located in different parts of the graph, yet positional information is underused in previous research.

Second, most approaches adopted shallow GNNs in propagating messages through local (mostly 2-hop) neighbors, causing the structure-awareness to be local. Shallow GNNs are popular because of the over-squashing effect \cite{alon2020bottleneck} and the over-smoothing issues \cite{liu2020towards,xu2021optimization} that occur when the number of layers increases.

Third, several works combining GNNs with CNNs were proposed to solve segmentation problems \cite{juarez2019joint,tan2021sgnet,shin2019deep}, where they operated on branch points using dense features from a segmentation network. Although these works can be applied to airway labeling problem, we argue that this is counter-intuitive and inefficient, because airway labeling intrinsically is a classification problem. In addition, using segmentation methods for labeling airways cannot guarantee that a branch has all its voxels assigned with the same label.

This paper offers an efficient airway labeling method by combining CNNs and GNNs. Nodes in our graph are branches, represented by CNN features. Nearby branches in our graph are connected following the exact tree topology. A GNN enriches the CNN branch features. Working at the branch level reduces computational cost as the number of nodes and edges decreases dramatically compared to branch points methods.

Our method also explores the possibilities of using deeper architectures with skip connections to reduce the over-smoothing effect \cite{chen2020simple}. We use a graph attention network \cite{velivckovic2017graph} as the backbone of our method because it is known for being less susceptible to over-squashing issues \cite{alon2020bottleneck}. 
Our method captures structural relationships between connected branches using the proposed graph neural network. The structure-awareness, also referred to as combinatorial property in a related work \cite{feragen2012}, is essential for airway labeling because airway trees or sub-trees may appear similar at individual branches yet are combinatorially different. In addition, we encode node positions as their rescaled shortest path lengths to a set of anchor nodes, and such positional encodings are used as additional features to improve the representation learning in GNN. 
We refer to our proposed method as a structure- and position-aware graph neural network (SPGNN).

Our key contributions are: 1) SPGNN introduces a novel idea of extracting structural and positional information from airway trees in representing branches for airway anatomical labeling. Our method achieves 91.18\% overall branch classification accuracy on a challenging dataset of COPD subjects, including all GOLD stages; 2) The SPGNN method is generic and can be extended to other tree branches classification problems such as labeling vessel trees. 3) Our approach is memory and computationally efficient. It requires only an NVIDIA GTX1080 GPU with 8GB memory to train and takes around 17 seconds to label an entire segmented airway tree.

\section{Data}\label{sec:data}

\begin{table}
  \caption{The distribution of GOLD stages in the main and secondary data collection. Gold stages 0-4 defined in \cite{regan2011COPDGene}. No PFT: spirometry data not available; PRISM:Preserved Ratio Impaired spirometry \cite{wan2014epidemiology} in the COPD data set.}
  \centering
  \begin{tabular}{lll}
    \toprule
    \multirow{2}{*}{GOLD stages}   & \multicolumn{2}{c}{\# subjects} \\
    \cline{2-3}
    & main  & secondary \\
    \midrule
    GOLD0     & 31  &  6  \\
    GOLD1     & 33  & 10  \\
    GOLD2     & 32  & 6   \\
    GOLD3     & 33  & 7   \\
    GOLD4     & 27  & 5   \\
    Non Spirometry  & 1 & 0 \\
    Non Smoking     & 32 & 4     \\
    PRISm     & 31  & 2  \\
    \bottomrule
     Total     & 220  & 40    \\
    \bottomrule
 \end{tabular}
 \label{tab:datameta}
\end{table}

We obtained chest CT scans from the COPDGene study \cite{regan2011COPDGene}, which is a clinical trial with data from 21 imaging centers in the United States. In total, COPDGene enrolled 10,000 subjects, and each subject underwent both inspiration and expiration chest CT. Data from COPDGene is publicly available and can be retrieved after submitting an ancillary study proposal (ANC-337 was used for this work).

We randomly selected 220 subjects from the COPDGene study as our main data collection for cross-validating our algorithm. The other 40 randomly selected subjects from the COPDGene study were used as the secondary data collection for a reader study. The main and the secondary data collection included subjects with various COPD severity stages. See Table \ref{tab:datameta} for the distribution of selected subjects in terms of their COPD gold stages. We utilized only inspiration CT scans from the first visit, one scan per subject. Among the selected inspiration CT scans, slice thicknesses are ranged from 0.625-0.9mm and pixel spacing from 0.478-1.0mm. In this work, CT scans were only used to extract the airway trees. The extracted airway trees were stored in label-map images and were used as the input to our algorithms.  

\subsection{Reference Standard}\label{sec:reference}
Since the first several generations of the human airway tree have a relatively similar topology across subjects, we focused on labeling the 18 segmental airways: 8 from the left lung (LB1+2, LB3, LB4, LB5, LB6, LB7+8, LB9, and LB10) and 10 from the right lung (RB1-10), following the anatomical labeling scheme from \cite{tschirren2005matching}. 

Given an inspiration chest CT scan, an airway tree and branch segments were extracted using the front propagation segmentation method \cite{van2008robust}, where voxels of each branch were assigned a random unique integer value increasing from 1 at the trachea to a larger value in its descendants. The generated airway segmentations were visually inspected by trained analysts and manual editing was made to ensure that all segmentations included airways up until at least the segmental level. We use these manually-edited segmentation maps as the input to our labeling algorithms. Next, based on their anatomical knowledge of the airway tree, analysts clicked on segmental branches to assign anatomical labels. The resulting manual airway labeling reference was a segmentation label map where 18 segmental branch segments were assigned with unique values.

All analysts have a medical background and have received extensive training in the segmentation and anatomical labeling of airways in CT imaging. The analysts could consult a radiologist in cases of doubt during the annotation process.

\section{Methods}

\begin{figure*}
\centering
 \includegraphics[width=0.9\textwidth]{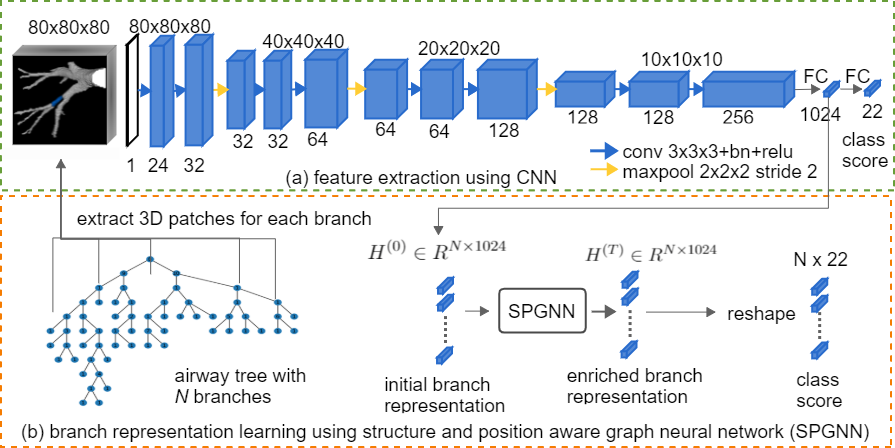}
\caption{The overview of our proposed airway labeling framework. Branch features extracted from the CNN ($a$) are enriched using the proposed SPGNN network ($b$). } 
\label{fig:framework}
\end{figure*}
\subsection{Airway Labeling Framework}
We formulate airway anatomical labeling as a branch classification problem. For training, we add the trachea and two main bronchi as additional target labels to provide more information regarding the airway tree hierarchy. Therefore, the classification targets for training are 22 classes, including 18 segmental airways, trachea, two main bronchi, and one additional class to represent all other airway branches. The overview of our branch classification framework is illustrated in Fig.~\ref{fig:framework}. To ensure the same spatial scale, we first resample the airway segmentation maps to a fixed voxel spacing (0.625mm, 0.625mm, and 0.5mm in sagittal, coronal, and axial views). We then use the resampled segmentation map to build a graph by considering each branch a node and making connections if two branches share a common boundary in the segmentation map. We first train a CNN to predict branch target labels. The branch features are then extracted as the features before the classification head in the CNN network. Meanwhile, the predicted anatomical branches by the CNN are used as anchors to compute positional encodings. Next, the CNN branch features and computed positional encodings are enriched with a multi-layer GNN by iteratively exchanging and gathering information between neighboring branches. Finally, branch features generated by the GNN are fed into a classification head to produce class probabilities. The following subsections describe this in detail.

\subsubsection{Branch feature extraction using CNN}\label{sec:CNN}
We train our CNN using the input of 3D patches cropped around each branch. To do so, we first skeletonize the airway segmentation using a thinning algorithm \cite{lee1994building}. The branch center is defined as the center of the skeletonized branch. At the branch center, we crop a 3D patch of size 80$\times$80$\times$80 voxels. 

In the 3D patch, voxels inside the branch of the interest at the patch center are set to 0.9. Voxels in other branches in the patch are set to 0.5, and the background is set to 0. We ignore the CT Hounsfield values to prioritize shape and connectivity information in branch representation learning.

Our CNN has three downsampling blocks. Each consists of two consecutive 3$\times$3$\times$3 convolutions and one 2$\times$2$\times$2 max-pooling with stride 2. After the down-sampling blocks we apply two more convolutions with kernel size 3 to double the number of filters up to 256 before flattening into 1024-dimensional feature vectors. These feature vectors are reshaped into 22-dimensional class vectors using a single linear layer (classification head) with a softmax activation function to produce class probabilities. We train this CNN using cross-entropy loss. Architectural details are provided in Figure \ref{fig:framework}. Branch features are the 1024-dimensional feature vectors before the classification head. The CNN branch features are fed to the GNN as the input for training. We report the classification performance using only CNN features as our baseline.

\subsubsection{Labeling airways using branch classification prediction}\label{sec:branch_prediction}
At test time, to label a segmented airway tree with $N$ branches, we run CNN predictions to generate a class probabilities matrix $C \in R^{N\times 22}$. The predicted branch of a target label at the column $s$ is indicated by the row index $i$ that maximizes probabilities at the column $s$, $i = argmax(C[, ..s.., ])$. We only apply this computation on columns corresponding to the 18 segmental branches. If the same branch is matched more than once, we assign the branch to the most confident label. Note that the total number of predicted labels may be less than 18 because there may also be missing anatomical branches in reference trees. 

\subsubsection{Enriching branch representation using GNN}\label{sec:GNN}
The proposed GNN network intends to capture each branch's structure (connectivity to other branches) and position information (location in the graph), referred to as the structure and position-aware graph neural network (SPGNN). The SPGNN network sequentially applies $l$ layers. Each layer passes and gathers messages within its 1-hop neighbors. As a result, stacking $l$ layers allows branches in the network to reach their $l$-hop neighbors. In such a way, each branch is represented by its features and features from its $l$-hop neighbors. We use two input features to the SPGNN network: the learned CNN branch features, focusing on branch visual characteristics and branch positional encodings.\par
The positional encoding serves the same purpose as the geometric prior in relational convolution neural networks \cite{hu2019local,xie2020relational} and position encodings in Transformers \cite{vaswani2017attention,dosovitskiy2020image}. Positional encoding aims to introduce a coordinate system into the learning process such that the orders and distances between objects are embedded in the learned representation. 
We introduce the concept of anchors to encode positional information for each branch, inspired by \cite{you2019position}. We use 18 segmental branches, the trachea, and two main bronchi predicted by the CNN as the initial anchors. The leaf branch rooted from a predicted segmental branch is also added to the set of anchors. We only select the leaf branch farthest from the predicted segmental branch among multiple leaf branches. Consequently, the final anchor set includes 39 branches. Then the positional encoding of each node is the shortest path length to the 39 anchor branches in the airway tree. The shortest path lengths are rescaled to $\left[ 0\sim1\right]$ range by dividing the longest distance between any pair of branches in the tree. To have exactly 22 branches predicted from the CNN network, we slightly modify the branch prediction process (\ref{sec:branch_prediction}). Given the class probability matrix $C \in R^{N\times 22}$ by the CNN prediction, the probability of a given branch indexed at the $i$th row belonging to a target label indexed at $s$th column is the entry $C[i,s]$. We assign the target label indexed at $s$th column to the branch whose row-index maximizes the probabilities $i = argmax(C[, ..s.., ])$, in a leave-one-out fashion. Once a branch has been labeled, its corresponding row in $C$ is excluded for finding maximum indices in other classes. Assuming the CNN predictions are mostly accurate (results in Table \ref{tab:qresult}), the correctly-predicted branches can be used to provide canonical positional encodings because their locations are consistently defined (up, down, left, and right) according to the airway tree anatomy. Many existing positional encoding methods, i.e., Laplacian eigenvectors \cite{dwivedi2020benchmarking}, random-walks encodings \cite{dwivedi2021graph}, and encodings using random anchors \cite{you2019position} provide non-canonical positional encodings because these methods operate on arbitrary graphs. By adding leaf branches, the anchors distribute evenly in terms of the depth of the tree, preventing selecting anchors from only upper side of the tree. The anchors distribute evenly across lobes by adding predicted segmental branches. Meanwhile, the anchors follow a fixed ordering across different trees. In addition, our positional encoding is distance-sensitive. The Euclidean distance between positional encodings of two branches far apart on the graph must be large and small for two branches nearby. We point out that many existing methods on airway labeling or segmentation also use location information, i.e., the spatial position of the bronchial centerline points \cite{bulow2006point}, and voxels’ coordinates and their Euclidean distances to the carina \cite{tan2021sgnet}. However, the problem of using voxel coordinates to present a branch location is that they are not topological, so two branches whose voxel coordinates are close in Euclidean space are not necessarily short in the path when traveling in the tree. Voxel coordinates are also sensitive to the image scale, to patient position and to variation in orientation.

For any branch, its positional encoding is a 39-dimensional vector, where each dimension indicates the rescaled shortest path length between the branch and one of the anchors, in the order of trachea, two main bronchi, 18 predicted segmental branches, and the 18 sub-segmental leaves according to the predictions of segmental airways. We denote an airway graph as $G = (B, E)$, where $B$ is a set of branches, $E$ is a set of edges connecting neighboring branches. Given an anchor set $S$ with $k$ anchors $S = \{s_{i}\}, i=1,...,k$, positional encoding for a branch $b$ can be written as $[d(b,s_{1}),d(b,s_{2}),..., d(b,s_{k})]$, where $d(\cdot, \cdot)$ is the rescaled shortest path length between two branches, where $[\cdot]$ is the concatenation operator.

Each layer $l$ in the SPGNN consists of two graph neural networks, denoted as $GNN_{hp}^{l}$ and $GNN_{p}^{l}$. At each layer $l$, the input to $GNN_{hp}^{l}$ is the concatenation between branch features $h^{l}$ and positional encodings $p^{l}$ from the previous layer, $h^{l} \in R^{N\times d_{h}^{l}}$ and $p^{l} \in R^{N\times d_{p}^{l}}$ where $N$ is the number of branches, $d_{h}^{l}$ is the dimension of $h^{l}$, and $d_{p}^{l}$ is the dimension of $p^{l}$.  The input to $GNN_{p}^{l}$ is the positional encoding $p^{l}$ from the previous layer. Therefore, $GNN_{p}^{l}$ focuses on learning positional encodings, whereas $GNN_{hp}^{l}$ relies on both branch features and positional encodings. At the first layer, the input feature to $GNN_{hp}^{1}$ is the concatenation of 1024-dimensional CNN feature and 39-dimensional positional encoding. The input to $GNN_{p}^{1}$ at the first layer is the 39-dimensional positional encoding. In SPGNN, layer-wise update can be formulated as:
\begin{align}
h^{l+1} =  \sigma(W_{hp}^{l}[h^{l},p^{l}] +  GNN_{hp}^{l}([h^{l},p^{l}]))\label{eq:spgnn_s}\\ 
p^{l+1} =  \sigma(W_{p}^{l}p^{l} +  GNN_{p}^{l}(p^{l}))\label{eq:spgnn_p}
\end{align}
where both $GNN_{hp}^{l}$ and $GNN_{p}^{l}$ are graph attention networks. $W_{hp}^{l}$ and $W_{p}^{l}$ are linear transformations to project the input features of the layer $l$ to be the same size as the output features from $GNN_{hp}^{l}$ and $GNN_{p}^{l}$, for enabling skip connections. The activation function $\sigma$ is the exponential Linear Unit. In SPGNN, we stack four layers. Given the 1063-dimensional (1024+39) input feature, the output features from the $GNN_{hp}$ are 256, 128, 64, and 1024 in dimensions from the first to the fourth layer. Given the 39-dimensional input positional encoding, the output encoding from the $GNN_{p}$ are 256, 128, and 64 in dimensions from the first to the third layer. $GNN_{p}$ has one layer less than $GNN_{hp}$ because we do not feed positional encodings into the classification head. In the end, the output of SPGNN is the 1024-dimensional branch feature from $GNN_{hp}$ at the fourth layer, which has the same size as the CNN branch feature. For predicting anatomical labels, a classification head using a single linear layer is to reshape the branch features into 22-dimensional class probabilities, the same as the classification head in the CNN method.\par
We adopt graph attention networks as the backbone in SPGNN for the following reasons. First, graph attention networks selectively attend over neighbors (anisotropic) for each branch via a self-attention mechanism which has proven to be useful for many machine learning tasks \cite{vaswani2017attention,dosovitskiy2020image}. Second, graph attention networks are non-spectral methods and have shown superior performance in inductive learning benchmarks \cite{velivckovic2017graph}, including tasks such as airway labeling where the model has to generalize to completely unseen graphs. Spectral methods such as GCNs and chebNet \cite{hammond2011wavelets} are intrinsically transductive because their exact solutions depends on the full graph Laplacian to be known during training. Third, graph attention networks are less susceptible to the over-squashing effect \cite{alon2020bottleneck}. Additionally, we add skip connection in SPGNN in between layers to alleviate over-smoothing issues, following the reasoning in \cite{xu2021optimization}.

For each branch (node) $b$, the graph attention network updates the node feature $h_{b}^{l}$ at layer $l$ using its neighbors $N_{b}$ as:
\begin{equation}\label{eq:gat}
\begin{array}{l}
h_{b}^{l+1} =  \sigma(\sum_{j\in N(b)} \alpha_{bj}^{l}W_{a}^{l}h_{j}^{l})\\
\alpha_{bj}^{l} = \frac{exp(\sigma(W_{r}^{l}[W_{g}^{l}h_{b}^{l},W_{g}^{l}h_{j}^{l}]))}{\sum_{k \in N_{b}}exp(\sigma(W_{r}^{l}[W_{g}^{l}h_{b}^{l},W_{g}^{l}h_{k}^{l}]))},\\
\end{array}
\end{equation}
where $\alpha_{bj}^{l}$ is the attention weight between branch $b$ and $j$ at layer $l$. $[\cdot]$ is the concatenation operator. $W_{a}^{l}$, $W_{r}^{l}$, and $W_{g}^{l}$ are linear transformations with $W_{a}^{l},W_{g}^{l}\in R^{d^{l}\times d^{l+1}},W_{r}^{l} \in R^{2d^{l+1} \times 1}$ where the input and output dimension of the layer $l$ is $d^{l},d^{l+1}$. $W_{r}^{l}$ projects the concatenated features into a scalar as the attention weight. The attention weights are normalized using the softmax function over all neighboring pairs. We add self-connections for nodes when training graph neural networks in our experiments.\par
The SPGNN without positional encodings can be formulated as:
\begin{equation}\label{eq:gnn}
h^{l+1} =  \sigma(W_{h}^{l}h^{l} + GNN_{h}^{l}(h^{l})),
\end{equation}
where $GNN_{h}^{l}$ is a graph attention network. Therefore, the SPGNN without positional encodings is equivalent to a graph attention network with skip connections. We refer to this network as GATS and compare this with SPGNN in order to see if there is a  benefit of using positional encodings.

\section{Results}
\begin{table*}[t]
\caption{Branch Classification Accuracy (ACC(\%)) and Topological Distance (TD) of the CNN, GATS, and the proposed SPGNN methods (in mean $\pm$ standard deviation). The overall branch classification accuracy is the accuracy of all target labels on average. Multiply accumulate operations (MACs) and the number of parameters are shown as measures of computational complexity. Testing time consumption indicates the run-time efficiency. The overall topological distance is the average of TD on all target labels. Boldface denotes the best result.} \label{tab:qresult}
\centering
\begin{minipage}[t]{\linewidth}
\centering
\begin{tabular}{l|l|l|l|l|l|l|l|l|l|l}
\hline
Method&Metric&LB1+2&LB3&LB4&LB5&LB6&LB7+8&LB9&LB10&RB1\\
\hline
\multirow{2}{*}{CNN}&ACC(\%)&\makecell{76.81}&\makecell{75.45}&\makecell{81.36}&\makecell{80.00}&\makecell{98.63}&\makecell{92.72}&\makecell{84.09}&\makecell{72.72}&\makecell{81.36}\\\cline{2-11}
&TD&\makecell{\textbf{1.36$\pm$0.59}}&\makecell{2.75$\pm$1.37}&\makecell{3.39$\pm$2.32}&\makecell{3.31$\pm$3.59}&\textbf{\makecell{2.00$\pm$0.81}}&\makecell{2.40$\pm$2.75}&\makecell{2.97$\pm$1.69}&\makecell{1.22$\pm$0.55}&\makecell{2.34$\pm$1.92}\\ 
\hline
\multirow{2}{*}{GATS}&ACC(\%)&\makecell{80.90}&\makecell{85.00}&\makecell{\textbf{88.63}}&\makecell{91.36}&\textbf{\makecell{99.54}}&\makecell{94.54}&\makecell{88.18}&\makecell{81.36}&\makecell{85.45} \\\cline{2-11}
&TD&\makecell{1.40$\pm$0.57}&\makecell{2.45$\pm$1.04}&\textbf{\makecell{2.28$\pm$1.07}}&\makecell{1.36$\pm$1.56}&\makecell{3.00$\pm$0.00}&\makecell{\textbf{1.66$\pm$1.02}}&\makecell{3.07$\pm$1.77}&\makecell{1.27$\pm$0.49}&\makecell{2.15$\pm$1.48}\\
\hline
\multirow{2}{*}{SPGNN}&ACC(\%)&\makecell{\textbf{82.27}}&\makecell{\textbf{87.27}}&\makecell{88.63}&\makecell{\textbf{92.27}}&\makecell{99.09}&\makecell{\textbf{95.45}}&\makecell{\textbf{91.36}}&\textbf{\makecell{83.18}}&\textbf{\makecell{87.72}} \\\cline{2-11}
&TD&\makecell{1.38$\pm$0.48}&\makecell{\textbf{2.42$\pm$0.56}}&\makecell{2.36$\pm$0.93}&\textbf{\makecell{1.05$\pm$0.23}}&\makecell{2.00$\pm$1.00}&\makecell{1.90$\pm$1.04}&\makecell{\textbf{2.57$\pm$1.09}}&\makecell{\textbf{1.16$\pm$0.44}}&\makecell{\textbf{1.85$\pm$1.29}}\\
\hline
\end{tabular}

\end{minipage}
\begin{minipage}[t]{\linewidth}
    \centering
\begin{tabular}{l|l|l|l|l|l|l|l|l|l|l}
\hline
&Metric&RB2&RB3&RB4&RB5&RB6&RB7&RB8&RB9&RB10\\
\hline
\multirow{2}{*}{CNN}&ACC(\%)&\makecell{87.72}&\makecell{83.18}&\makecell{79.09}&\makecell{85.00}&\makecell{95.90}&\makecell{91.36}&\makecell{90.00}&\makecell{80.90}&\makecell{72.72}\\\cline{2-11}
&TD&\makecell{2.44$\pm$2.06}&\makecell{1.78$\pm$2.34}&\makecell{1.95$\pm$2.61}&\makecell{3.59$\pm$3.20}&\makecell{2.88$\pm$2.84}&\makecell{2.50$\pm$1.32}&\makecell{2.52$\pm$1.56}&\makecell{2.69$\pm$1.37}&\makecell{1.41$\pm$0.73}\\ 
\hline
\multirow{2}{*}{GATS}&ACC(\%)&\makecell{90.45}&\makecell{87.72}&\makecell{88.18}&\makecell{94.09}&\makecell{\textbf{99.54}}&\makecell{95.45}&\makecell{93.63}&\makecell{90.45}&\makecell{82.72} \\\cline{2-11}
&TD&\textbf{\makecell{1.76$\pm$1.23}}&\makecell{1.85$\pm$2.60}&\makecell{1.96$\pm$1.97}&\makecell{2.25$\pm$1.08}&\makecell{\textbf{1.00$\pm$0.00}}&\makecell{2.80$\pm$0.74}&\makecell{\textbf{2.00$\pm$1.13}}&\makecell{2.80$\pm$1.59}&\makecell{\textbf{1.26$\pm$0.59}}\\
\hline
\multirow{2}{*}{SPGNN}&ACC(\%)&\makecell{\textbf{91.36}}&\makecell{\textbf{90.90}}&\textbf{\makecell{90.00}}&\makecell{\textbf{95.00}}&\makecell{98.63}&\makecell{\textbf{95.90}}&\makecell{\textbf{96.36}}&\textbf{\makecell{92.72}}&\textbf{\makecell{83.18}} \\\cline{2-11}
&TD&\makecell{1.78$\pm$0.89}&\makecell{\textbf{1.40$\pm$0.66}}&\makecell{\textbf{1.36$\pm$0.56}}&\textbf{\makecell{2.00$\pm$1.18}}&\makecell{1.00$\pm$0.00}&\makecell{\textbf{2.11$\pm$0.73}}&\makecell{2.50$\pm$0.95}&\makecell{\textbf{2.25$\pm$1.08}}&\makecell{1.29$\pm$0.45}\\
\hline
\end{tabular}
\end{minipage}
\begin{minipage}[t]{\linewidth}
    \centering
    \vspace*{2mm}
{\begin{tabular}{l|l|l|l|l|l}
\hline
Method overall&ACC(\%)&TD&MACs&\#param&testing time (second)\\
\hline
CNN&83.83$\pm$7.37&2.41$\pm$0.67&\textbf{6.42G}&\textbf{67.49M}&\textbf{14.25$\pm$9.65}\\
\hline
GATS&89.84$\pm$5.44&2.02$\pm$0.61&6.62G&69.52M&16.12$\pm$8.69\\
\hline
SPGNN&\textbf{91.18$\pm$4.97}&\textbf{1.80$\pm$0.50}&6.67G&70.09M&16.98$\pm$9.79\\
\hline
\end{tabular}
}
\end{minipage}
\end{table*}

\subsection{Experimental details}\label{sec:exp_setting}
All experiments were carried out on a machine with an NVIDIA GTX1080 with 8 GB GPU memory. All methods were evaluated by running 5-fold cross-validation on the main data collection with 220 airways (details in \ref{sec:data}). Our airway labeling method consists of two sequential steps: training CNN network and training SPGNN network. Each network has a linear layer as its classification head. Both networks were optimized using stochastic gradient descent with momentum 0.9 and weighted cross-entropy loss. All methods were implemented using Python 3.8 and Pytorch 1.7.1 library \cite{Pasz19}. Graph neural networks were implemented using the DGL graph computing library \cite{wang2019deep} version 0.6.1. Model parameters were initialized according to \cite{He15}. The initial learning rate was set to $5^{-4}$. Training on each fold stopped at epoch 150. Both models took nearly 48 hours to train per fold. For each cross-validation split, we trained using the training folds, and applied the trained networks to the test fold. Finally, we merged the results on the test folds of each split for evaluation. 

We used two evaluation metrics for measuring the overall classification performance. The branch classification accuracy (ACC) for each target label is the number of correct predictions divided by the total number of branches for that label on the dataset (at most 220 branches for each label). The overall branch classification accuracy is the accuracy of all target labels on average. Given the airway tree graph, the topological distance (TD) is the shortest path length between the predicted branch and the target branch. TD measurements were only computed on mislabeled airway branches because all correctly labeled branches have a topological distance of 0. The overall topological distance is the average of TD on all target labels. We also report computational complexity as the number of multiply-accumulate operations (MACs) and the number of network parameters. Runtime efficiency was measured by the test processing time per case on average.

\subsection{Quantitative results}\label{sec:main_results}
Three branch classification methods are compared: 1) the baseline using branch features from the CNN; 2) the GATS (Eq.~\ref{eq:gnn}), equivalent to the SPGNN without learnable positional encodings; 3) the SPGNN. The GATS method is only structure-aware, whereas SPGNN is both structure and position-aware. 

As shown in Table \ref{tab:qresult}, the GATS method substantially outperforms the baseline by learning structural information within four-hop neighbors, from 83.83\% to 89.84\% in ACC and from 2.41 TD to 2.02 TD. Adding positional encodings in SPGNN further improves the results, reaching 91.18\% ACC and 1.80 TD. In terms of model complexity and runtime efficiency, the baseline method has only slightly fewer parameters, MACs, and less time consumption on processing a test scan than those in GATS and SPGNN because both GATS and SPGNN depend on the feature extraction using the CNN network. The difference between GATS and SPGNN is trivial in model complexity and runtime efficiency. The SPGNN method can process a scan in roughly 17 seconds in an NVIDIA GTX1080 GPU with 8 GB memory. 

Regarding scores on individual branches, LB6 and RB6 achieve the highest ACC because these branches exhibit only minor anatomical variations. Branches with high anatomical variations such as LB1-4, LB10, and RB10 show lower ACC scores. It is also challenging to manually label these anatomical branches; we have noticed that human analysts may occasionally assign LB1 and LB2 (children of LB1+2) as LB1+2 when the actual LB1+2 is missing due to anatomical variation. In terms of ACC, graph-based approaches perform substantially better than the baseline in LB4, LB5, RB4, RB5, LB3, LB10, and RB10. This is because graph-based approaches can infer branch labels based on the predictions of nearby branches using the learned structural information. This knowledge propagation can result in an essential improvement in labeling branches with high anatomical variations such as LB10 and RB10 using predictions on nearby branches such as LB7+8 and RB7 with higher prediction accuracy. In terms of TD measurements, the SPGNN method outperforms other methods in most branches. The standard deviation of TD is much lower in the SPGNN than other methods in RB2, RB3, RB4, LB5, RB8, and RB9. This indicates that introducing positional encodings limits errors to nearby branches. 
A relatively large TD is seen in some CNN predictions, showing that the CNN is sensitive to shape variations due to the lack of structural information. Enlarging the receptive field using a deep CNN architecture or using a larger input size may add contextual information but will also non-linearly increase computational complexity. 


\subsubsection{Ablation study on architecture choices}\label{sec:ablation}
We conducted three ablation studies: the first for evaluating different GNN architectures in the airway labeling problem; the second for validating the number of layers needed for graph attention networks; the third for assessing the contribution of learnable positional encodings. Ablation studies were carried out using an identical experimental setup as for the main results (\ref{sec:main_results}).

The GNN architectures we compare are the graph attention network (GAT) \cite{velivckovic2017graph}, graph isomorphic network (GIN) \cite{xu2018powerful}, GraphSage \cite{hamilton2017inductive} (SAGE), and Graph convolution network (GCN) \cite{kipf2016semi} because they have shown superior performance on many graph neural network benchmarks. In the ablation study, GAT denotes the vanilla graph attention network without skip connections, different from the GATS method that uses skip connections (Eq.~\ref{eq:gnn}). Architectures in comparison have four layers, and each takes 1024-dimensional CNN branch features as the input and produces equal-sized features as the output. At each layer, branch features are updated using neighboring information. Branch features are projected to 256, 128, 64, and 1024 dimensions from the first to the fourth layer, same as settings in SPGNN (\ref{sec:GNN}). We do not use positional encodings in ablation studies because we focus on analyzing the benefits of using different architectures for airway labeling.

In SAGE, sampling neighbors is unnecessary for airway trees because node degrees are relatively small compared to graphs in large-scale benchmarks. The aggregation function in SAGE is the max pooling operator with a linear layer \cite{hamilton2017inductive}. In GIN, average pooling is used in the aggregation function. As shown in Table \ref{tab:abla_arch} (a), GAT achieves the best performance in both metrics, reaching 89.59\% overall branch classification accuracy and 2.05 topological distance. We noticed that a four-layer GCN converges to degenerated results in ACC due to the known over-smoothing effect \cite{li2018deeper}.

Interestingly, over-smoothed features mainly cause classification errors within local neighbors as it shown by a relatively small TD (1.78). GIN, SAGE, and GAT substantially outperform the CNN baseline (ACC: 83.83\%, TD:2.41), which demonstrates the importance of using neighboring information in learning branch features. Graph attention networks perform better than GIN and SAGE, indicating that weighted averaging via attention offers better messaging aggregation than the techniques in GIN and SAGE. 

The second ablation study is to quantify the performance gap of using different layers in GAT. We compare results when stacking two, four, and seven layers. We use the same experimental settings for training the network with two and four layers as mentioned in \ref{sec:exp_setting}. For training the network stacking seven layers, we lower the learning rate to $1^{-5}$ to avoid exceptional large gradients, and therefore we allowed training to reach 250 epochs. The projected feature dimensions are 256 and 1024 for the network stacking two layers. For the network stacking seven layers, projected feature dimensions are 256, 128, 64, 64, 64, 64, and 1024 from the first to seventh layer. From Table~\ref{tab:abla_arch} (b), with or without skip connection, using 2-hop neighbors already provides substantially improved results compared to the CNN baseline (ACC: 83.83\%, TD:2.41). Stacking seven layers achieves a slightly better performance than the network stacking four layers, in the method with skip connection. We also show skip connections can help to alleviate over-smoothing in the GAT, from the performance drop of using the seven layer network without skip connections. 

The final ablation study assesses the contribution of using learnable position encodings. In Transformers \cite{vaswani2017attention}, positional encodings are concatenated with the word embeddings as the input to the learning process without being involved in the layer-wise update, referred to as non-learnable positional encodings (NLPE). Similar to \cite{dwivedi2021graph}, in our work, we iteratively update positional encodings using a dedicated GNN such that positional encodings can be adjusted to the graph structure at hand. The result using learnable position encodings is denoted as PE. Results in Table~\ref{tab:abla_arch} (c) show that making positional encodings learnable only contributes to minor improvements in branch classification performance. However, allowing positional encodings to be learned makes our method data-driven, thus can potentially improve the generalization capability of our method.

\begin{table}
 \caption{
  Ablation study on architecture choices. We compare GCN (\cite{kipf2016semi}), GIN (\cite{xu2018powerful}), SAGE (\cite{hamilton2017inductive}), and GAT (\cite{velivckovic2017graph}) when stacking fours layers for each architecture in (a). For GAT, we compare results when stacking two, four, and seven layers in (b). The benefit of making positional encoding learnable is shown in (c). The best result shown in bold.}
 \centering
\caption*{(a) GNN variants}
 \begin{tabular}{lll}
  \toprule
  Method    & ACC (\%) & TD  \\
  \midrule
  GCN     & 68.35$\pm$10.95 & \textbf{1.78$\pm$0.37}     \\
  GIN     & 88.07$\pm$7.15 & 2.03$\pm$0.68     \\
  SAGE    & 88.81$\pm$5.80 & 2.25$\pm$0.54      \\
  GAT     & \textbf{89.59$\pm$5.76} & 2.05$\pm$0.67     \\
  \bottomrule
\end{tabular}
\vspace*{0.5 cm}
\caption*{(b) \#stacked layers}
\begin{tabular}{lll}
  \toprule
  \#layers & ACC (\%) & TD \\
  \midrule
  2 & 87.26$\pm$6.78 & 2.11$\pm$0.59     \\
  2+skip connection & 87.90$\pm$6.89 & 2.00$\pm$0.62     \\
  4 & 89.59$\pm$5.76 & 2.05$\pm$0.67     \\
  4+skip connection & 89.84$\pm$5.44 & 2.02$\pm$0.61     \\
  7 & 82.19$\pm$7.18& 2.18$\pm$0.79      \\
  7+skip connection & \textbf{90.30$\pm$5.30} & \textbf{2.09$\pm$0.76}      \\ 
\bottomrule
\end{tabular}
\vspace*{0.5 cm}
\caption*{(c) learnable vs non-learnable positional encodings}
\begin{tabular}{lll}
  \toprule
  Method & ACC (\%) & TD \\
  \midrule
  NLPE & 90.98$\pm$5.32 & 1.97$\pm$0.82     \\
  PE & \textbf{91.18$\pm$4.97} & \textbf{1.80$\pm$0.50}      \\
\bottomrule
\end{tabular}
\label{tab:abla_arch}
\end{table}

\subsection{Comparison with human readers}
We invited two analysts to participate in a reader study on the secondary data collection (\ref{sec:data}) in which the analysts had not yet seen airways. The secondary data collection included 40 subjects with various COPD severities. Two analysts independently annotated anatomical labels given segmented airway trees. The same segmented trees were presented as the input to the proposed SPGNN algorithm. We computed a linearly weighted kappa with 95\% confidence interval in pair groups among the two analysts and the proposed SPGNN algorithm using the R software package (version 3.6.2; R Foundation for Statistical Computing, Vienna, Austria). Reader agreement was categorized as slight, fair, moderate, good, or excellent based on $k$ values of 0.20 or less, 0.21–0.40, 0.41–0.60, 0.61–0.80, and 0.81 or higher, respectively. Table~\ref{tab:kappa} demonstrates that the agreement between observers is excellent, same as the agreement between any of the analysts and the proposed algorithm (SPGNN), although the agreement between observers is slightly higher.

\begin{table}
  \caption{Reader study using linearly weighted kappa scores (95\% confidence interval) between paired groups among the two analysts and the proposed SPGNN algorithm.}
  \centering
  \begin{tabular}{ll}
    Paired Groups     & Kappa(95\%CI) \\
    \midrule 
   Observer 1 versus Observer 2     & 86.93 (76.26,97.60)      \\
   Observer 1 versus SPGNN     & 82.44 (71.78,93.10)     \\
   Observer 2 versus SPGNN     & 83.97 (73.30,94.65)     \\
    \bottomrule
 \end{tabular}
 \label{tab:kappa}
\end{table}

\subsection{Visualization of learned branch features}
\begin{figure*}
\minipage{0.32\textwidth}
  \includegraphics[width=\linewidth]{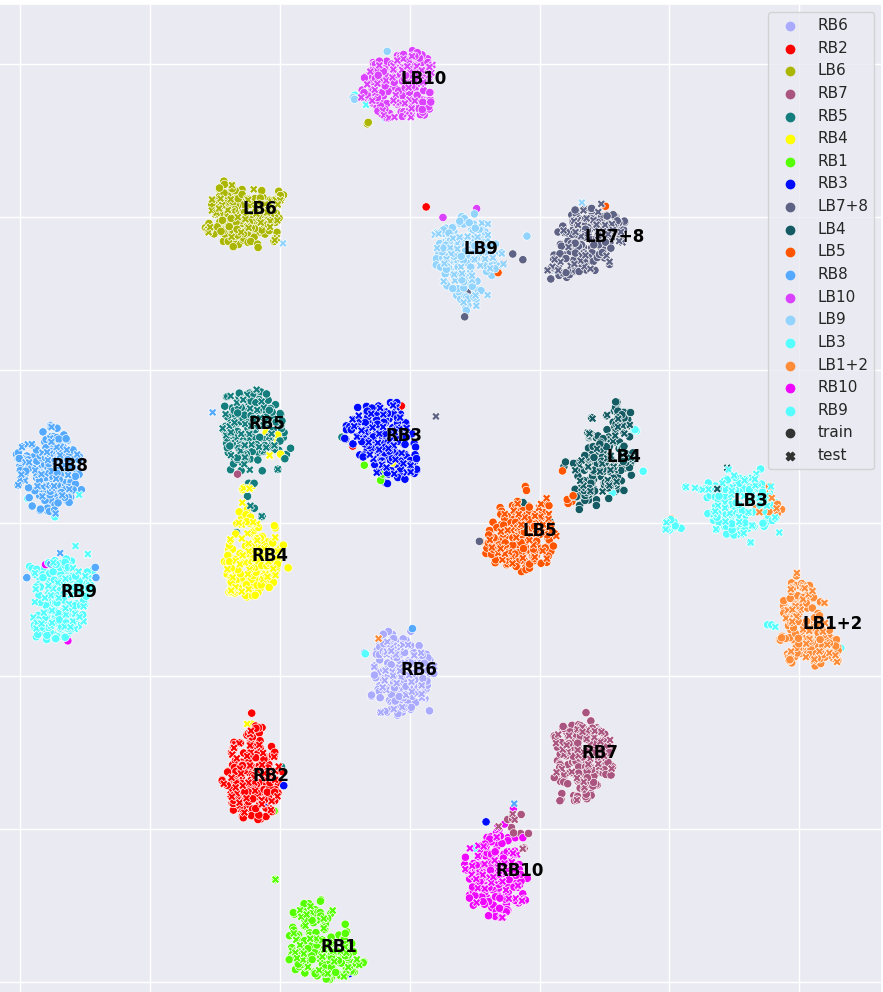}
  \captionsetup{labelformat=empty}
  \caption{CNN features}\label{fig:cnn_feat}
\endminipage\hfill
\minipage{0.32\textwidth}
  \includegraphics[width=\linewidth]{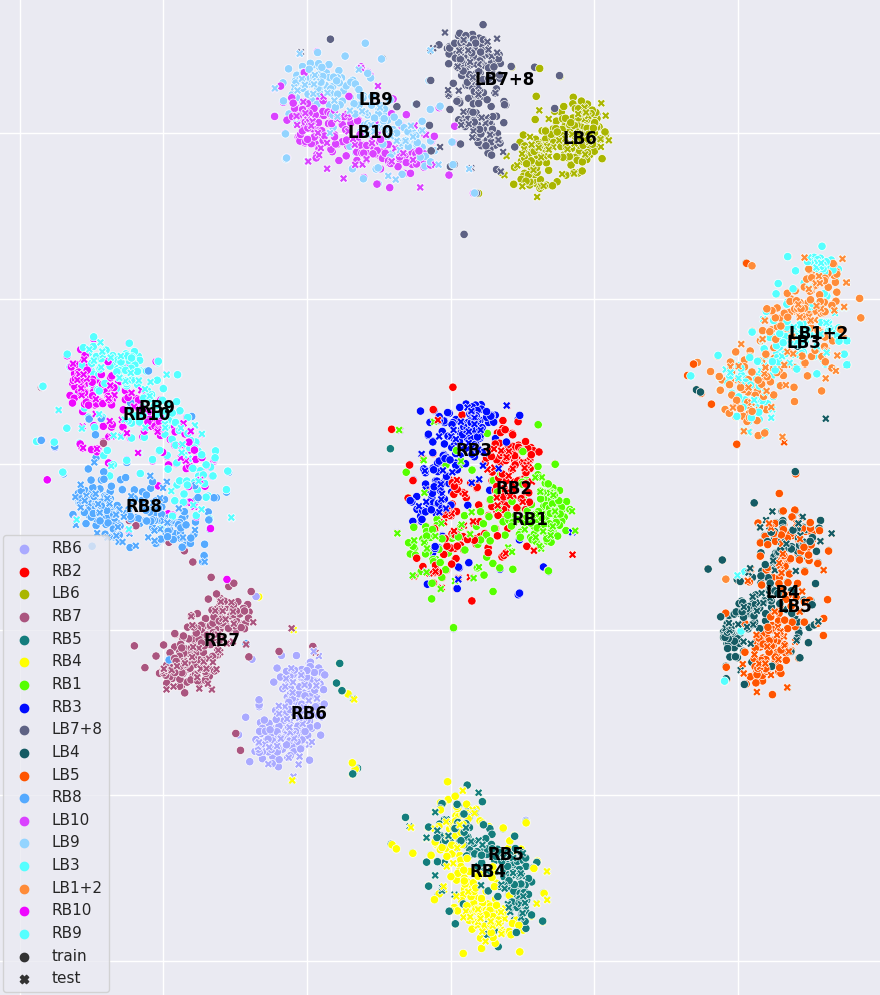}
  \captionsetup{labelformat=empty}
  \caption{Positional encodings}\label{fig:pos_feat}
\endminipage\hfill
\minipage{0.32\textwidth}%
  \includegraphics[width=\linewidth]{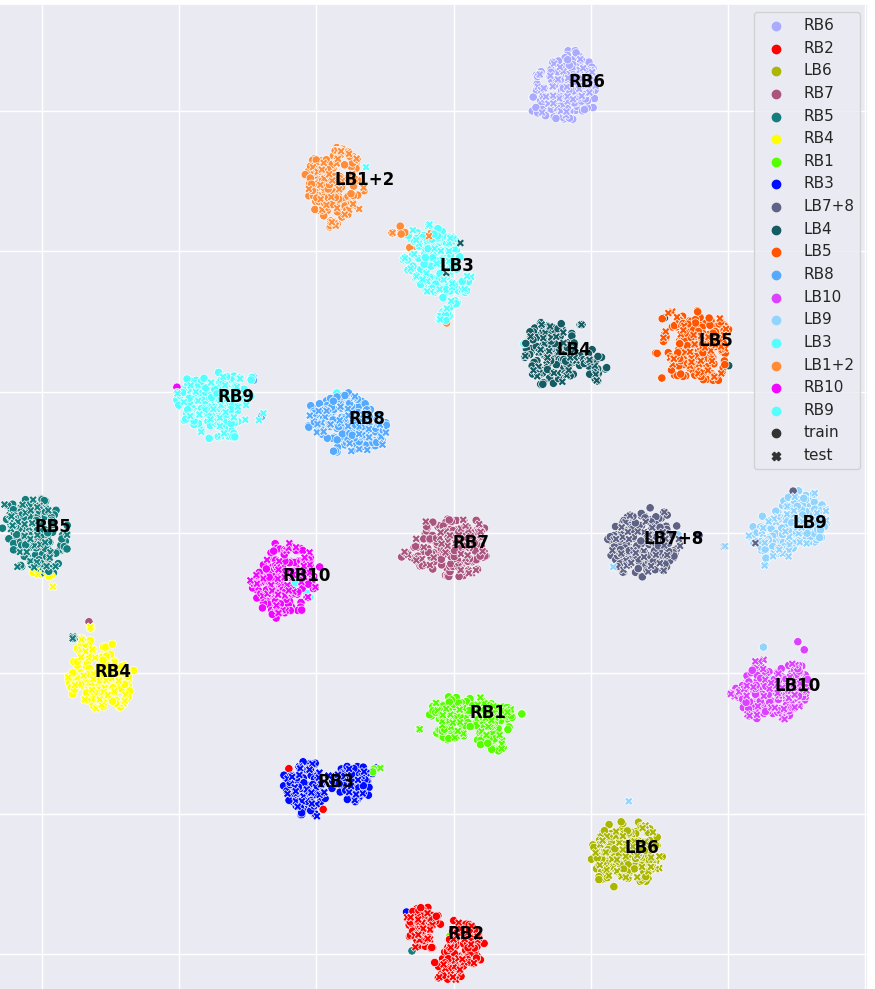}
  \captionsetup{labelformat=empty}
  \caption{SPGNN features}\label{fig:spgnn_feat}
\endminipage
\caption{t-SNE plots of the CNN features (left), positional encodings (middle) and SPGNN features (right) for all branches on 220 airways. Plots were generated by the trained model on one cross-validation split. Features from training or testing examples are marked by dot or cross. Features are colored according to their anatomical label, and anatomical names are annotated at the coordinate center among points of the same label.}
\label{fig:tsne}
\end{figure*}
In this section, we use the t-SNE plot \cite{van2008visualizing} to visualize learned branch features and positional encodings from the proposed SPGNN method in comparison with the learned branch features from the CNN method. Both CNN and SPGNN branch features are 1024-dimensional and learned positional encodings have 64 dimensions. We apply PCA to reduce 1024-dimensional branch features to 64 dimensions before applying t-SNE. We use perplexity 50 and max iterations 1000 for t-SNE optimization. We use the network trained on one cross-validation split (fold-0) to generate features on 220 airway trees in the main data collection. Features from training and testing split are visualized using different marker types (dot and cross). Features are labeled in different colors according to their anatomical label. The anatomical name is positioned at the coordinate center among points of the same label. 

The result is shown in Fig.~\ref{fig:tsne}. CNN branch features generally achieve good separation in feature space with regards to anatomical labels. However, there are confusions between features of nearby branches; for example, feature separations between RB4 and RB5,  between LB4 and LB5, and between LB9 and LB7+8 are much more visible in the SPGNN feature space, in comparison with those in the CNN feature space. The improved feature separation in SPGNN is because of the learned structural information. This phenomenon is consistent with quantitative results in Table \ref{tab:qresult} as a substantially improved branch classification accuracy for RB4, RB5, LB4, LB5, and LB9.

The positional encodings form a lobe-wise separation among branches. Therefore, the positional encodings could help to reduce misclassification across lobes, as shown in a reduced topological distance between SPGNN and other methods (Table~\ref{tab:qresult}). Nevertheless, positional encodings do not distinguish neighboring branches within the same lobe because neighboring branches have similar shortest path lengths to anchor nodes regardless of the selection of anchors. Additionally, we do not see a clear separation in feature space between training and test examples, indicating feature extraction process generalizes well between the train and the test split.

\subsection{Qualitative Results}
\begin{figure*}
\centering
  \includegraphics[width=0.8\linewidth]{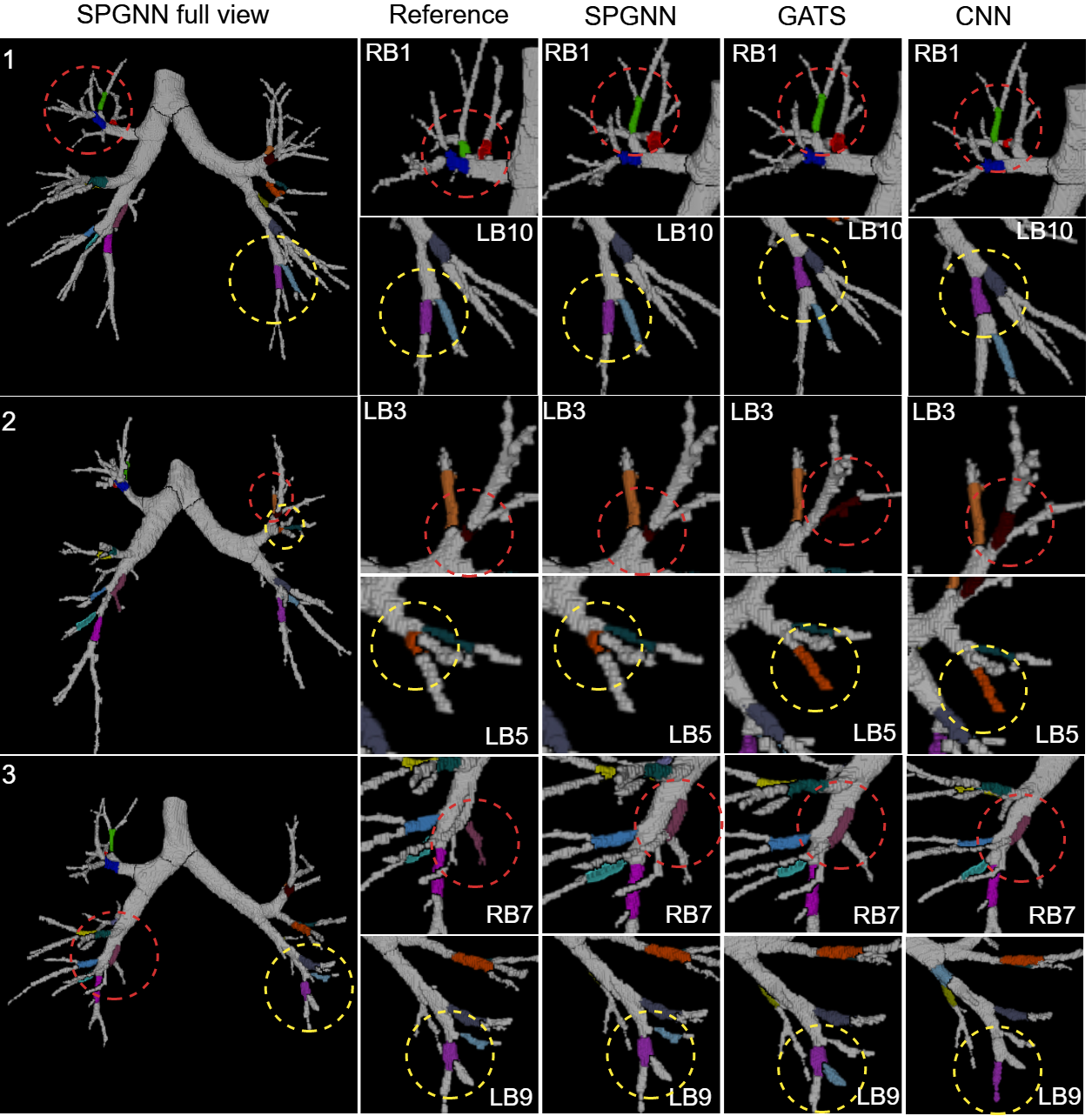}
\caption{Branch predictions for three representative cases in rows. In the first column, we show predictions from the SPGNN method in the full view. From the third column to the last, we show two close-up views highlighting the branch predictions generated by the proposed SPGNN, GATS, and the baseline CNN methods, respectively, according to the reference in the second column. 18 segments are in different colors as RB1 \textcolor{RB1}{\rule{.2cm}{.2cm}}, RB2 \textcolor{RB2}{\rule{.2cm}{.2cm}}, RB3 \textcolor{RB3}{\rule{.2cm}{.2cm}}, RB4 \textcolor{RB4}{\rule{.2cm}{.2cm}}, RB5 \textcolor{RB5}{\rule{.2cm}{.2cm}}, RB6 \textcolor{RB6}{\rule{.2cm}{.2cm}}, RB7 \textcolor{RB7}{\rule{.2cm}{.2cm}}, RB8 \textcolor{RB8}{\rule{.2cm}{.2cm}}, RB9 \textcolor{RB9}{\rule{.2cm}{.2cm}}, RB10 \textcolor{RB10}{\rule{.2cm}{.2cm}}, LB1+2 \textcolor{LB1+2}{\rule{.2cm}{.2cm}}, LB3 \textcolor{LB3}{\rule{.2cm}{.2cm}}, LB4 \textcolor{LB4}{\rule{.2cm}{.2cm}}, LB5 \textcolor{LB5}{\rule{.2cm}{.2cm}}, LB6 \textcolor{LB6}{\rule{.2cm}{.2cm}}, LB7+8 \textcolor{LB7+8}{\rule{.2cm}{.2cm}}, LB9 \textcolor{LB9}{\rule{.2cm}{.2cm}}, LB10 \textcolor{LB10}{\rule{.2cm}{.2cm}}}
\label{fig:quality_results}
\end{figure*}

In Fig.~\ref{fig:quality_results}, we show predictions of the SPGNN method by visualizing branch predictions in three airway trees in rows. The first column shows the full view of three airway trees where the proposed SPGNN method predicts the segmental labels. We zoom in on two branches for each case to demonstrate the reference annotation (2nd column), predictions from the SPGNN (3rd column), the GATS (4th column), and the baseline CNN methods (5th column). 

In the first case, the SPGNN, the GATS, and the CNN methods label RB1 as a sub-segmental branch rooted from RB1. This error may be caused by high anatomical variations in RB1, RB2, and RB3. There are cases where RB1, RB2, and RB3 trifurcate, while it is also common to have one bifurcation leading to RB1 before another bifurcation leading to RB2 and RB3. The LB10 is correctly labeled by the SPGNN method as a sibling of the LB9 but mislabeled as a sibling of the LB7+8 in the GATS and CNN methods. 

In the second case, the LB3 and the LB5 are mislabeled by GATS and CNN methods as one branch below, whereas the SPGNN makes accurate predictions. We noticed that the GATs and the CNN methods do not fully recognize sibling structures because they label both LB1+2 and LB4 correctly when mislabeling their siblings. On the other hand, the SPGNN labels LB3 and LB5 correctly, indicating that adding positional encodings in the SPGNN helps understanding sibling relationships. 

In the third case, the CNN, GATS, and SPGNN methods mislabel the RB7 as an unusual branch bifurcated before the RB7 from the right lower lobar bronchus. This additional split also causes the RB7 to change orientation relative to the right lower lobe bronchus. For the LB9, the CNN predicts the LB9 as one branch preceding the LB6, which causes a significant error in the TD metric. The GATS mislabels the LB9 as one branch succeeding the LB10, thus reducing the TD error. Meanwhile, the CNN method labels the LB10 as the branch below the LB10. However, this error does not reoccur in the GATS and the SPGNN results, indicating that structural information can reduce the classification error caused by the confusion in convolution features.  

\section{Discussion and Conclusion}

We have presented a method that formulates airway labeling as a branch classification problem, for which an accurate solution can be found by combining the power of CNNs and GNNs. We train a CNN to extract features for representing airway branches. These features are iteratively updated in a GNN by collecting the information from neighbors of each branch. The graph is built based on the airway tree connectivity. Furthermore, we leverage positional information in designing our GNN, where the position of each branch is encoded by its topological distance to a set of anchor branches. As a result, the learned features are structure- and position-aware, contributing to substantially improved branch classification results compared with methods using only convolution features or using only structure-aware GNNs.

By experimenting with various GNN architectures, we demonstrated that graph attention networks achieve better performance on our data set compared to other popular GNN architectures. By applying skip connections, we show that graph attention networks are resilient to the over-smoothing issues seen as performance drops in other experiments when stacking seven layers of the graph attention networks or stacking four layers of graph convolution networks without skip connection. However, the performance gained by using deeper architecture in GNN design is not substantial. 

In conclusion, we have shown that the proposed SPGNN achieves the top branch classification performance in our data set with only trivial computational overhead compared to other approaches in comparison. The proposed algorithm is publicly available at https://grand-challenge.org/algorithms/airway-anatomical-labeling/. Our method is generic and can be readily applied to other tree labeling problems that are ubiquitous in medical image analysis. We published our source code at https://github.com/DIAGNijmegen/spgnn.

\bibliography{ref.bib}

\end{document}